\documentclass[conference]{IEEEtran}
\IEEEoverridecommandlockouts
\usepackage{cite}
\usepackage{amsmath,amssymb,amsfonts}
\usepackage{algorithmic}
\usepackage{graphicx}
\usepackage{textcomp}
\usepackage[hidelinks]{hyperref}
\usepackage{subcaption}
\usepackage{xcolor}
\def\BibTeX{{\rm B\kern-.05em{\sc i\kern-.025em b}\kern-.08em
    T\kern-.1667em\lower.7ex\hbox{E}\kern-.125emX}}
\begin{document}
\bstctlcite{MyBSTcontrol}

\title{Accurate Gigapixel Crowd Counting by Iterative Zooming and Refinement}

\author{Arian~Bakhtiarnia,
        Qi~Zhang,
        and~Alexandros~Iosifidis
\thanks{Arian Bakhtiarnia, Qi Zhang and Alexandros Iosifidis are with DIGIT, the Department of Electrical and Computer Engineering, Aarhus University, Aarhus, Midtjylland, Denmark (e-mail: arianbakh@ece.au.dk; qz@ece.au.dk; ai@ece.au.dk).}
\thanks{This work was funded by the European Union’s Horizon 2020 research and innovation programme under grant agreement No 957337, and by the Danish Council for Independent Research under Grant No. 9131-00119B.
}}

\maketitle

\begin{abstract}
The increasing prevalence of gigapixel resolutions has presented new challenges for crowd counting. Such resolutions are far beyond the memory and computation limits of current GPUs, and available deep neural network architectures and training procedures are not designed for such massive inputs. Although several methods have been proposed to address these challenges, they are either limited to downsampling the input image to a small size, or borrowing from other gigapixel tasks, which are not tailored for crowd counting. In this paper, we propose a novel method called GigaZoom, which iteratively zooms into the densest areas of the image and refines coarser density maps with finer details. Through experiments, we show that GigaZoom obtains the state-of-the-art for gigapixel crowd counting and improves the accuracy of the next best method by 42\%.
\end{abstract}

\section{Introduction}
\label{sec:intro}

Crowd counting has many applications in video surveillance, social safety and crowd analysis and is an active area of research in the literature \cite{gao2020cnn}. Since most crowd counting applications and datasets use surveillance footage, the input to crowd counting models are high-resolution images, typically Full HD (1,920$\times$1,080 pixels) or even higher. Gigapixel resolutions can capture and process much more detail than previously possible, and are recently becoming more widespread \cite{bakhtiarnia2022efficient}. However, working with gigapixel resolutions presents several unique challenges. Modern high-end GPUs are not capable of fitting gigapixel images in memory or processing such high resolutions in reasonable time. Furthermore, the architectures of deep neural networks are not designed to receive such massive images as input.

Recently, several methods have been proposed for crowd counting on gigapixel images. However, these methods either use the simplest solution, which is to downsample the input gigapixel to a manageable resolution before processing, or borrow from gigapixel literature in other deep learning tasks. The issue with the latter approach is that gigapixel methods for other deep learning tasks such as object detection or cancer detection do not tackle unique challenges present in crowd counting, such as reliance on global information and sensitivity to perspective. On the other hand, the proposed method called \textit{GigaZoom} is tailored to crowd counting and is thus able to obtain significantly more accurate results compared to previous methods. GigaZoom works by iteratively zooming into the densest areas of the image and refining the coarser density map with finer details. Our code is publicly available\footnote{\url{https://gitlab.au.dk/maleci/gigazoom}}.

The rest of this paper is organized as follows. Section \ref{sec:related} summarizes the related work in crowd counting and gigapixel deep learning literature. Section \ref{sec:method} presents the proposed method. Section \ref{sec:experiments} describes the experimental setup and provides experimental results as well as ablation studies. Finally, section \ref{sec:conclusion} concludes the paper by summarizing contribution and results, and providing directions for future research.

\section{Related Work}
\label{sec:related}

\subsection{Crowd Counting}

The goal of crowd counting is to count the total number of people present in a given input image \cite{gao2020cnn}. The input to crowd counting models is an image or a video frame, and the output is a density map showing the crowd density at each location of the image. The values in the density map can be summed up to obtain a single number representing the total number of people in the image. Widely used crowd counting datasets contain high resolution images, for instance, images in Shanghai Tech Part A and Part B datasets \cite{zhang2016single} have average resolutions of 868$\times$589 and 1,024$\times$768 pixels, respectively, and images in the UCF-QNRF dataset \cite{Idrees_2018_ECCV} have an average resolution of 2902$\times$2013 pixels. However, these resolutions are much lower than gigapixel resolutions. At the time of this writing, PANDA \cite{Wang_2020_CVPR} is the only publicly available dataset for gigapixel crowd counting. PANDA contains 45 images with resolutions up to 26,908$\times$15,024 pixels taken from three different scenes: an airport terminal, a graduation ceremony, and a marathon. Images in the PANDA dataset are extremely densely populated with crowd sizes of up to 4,302 people, and ground truth annotations are available in the form of bounding boxes for each person's head. PANDA offers no predefined training or test splits.

Various crowd counting methods exist in the literature. CSRNet \cite{Li_2018_CVPR} uses the first ten layers of VGG-16 \cite{DBLP:journals/corr/SimonyanZ14a}, pre-trained on ImageNet \cite{5206848}, as a feature extractor, which is followed by six dilated convolution layers to produce the output density map. Gigapixel CSRNet \cite{cao2019gigapixel} utilizes CSRNet to process gigapixel images. During the training phase, CSRNet is trained on image patches of size 1,920$\times$1,200 pixels, taken across three scales: the original gigapixel image, as well as the gigapixel image downsampled to $\frac{1}{16}$ and $\frac{1}{64}$ of the original size. In the inference phase, image patches of the same size are passed on to the trained CSRNet in non-overlapping sliding windows to produce a density map for each scale. The three density maps are then averaged to obtain a single aggregated density map.

PromptMix \cite{bakhtiarnia2023promptmix} downsamples gigapixel images to 2,560$\times$1,440 pixels, then processes them using CSRNet. It improves the accuracy of CSRNet by mixing artificially generated data with real data during training. SASNet \cite{Song_Wang_Wang_Tai_Wang_Li_Wu_Ma_2021} is a high-performing crowd counting method on various popular datasets such as Shanghai Tech and UCF-QNRF. Similar to CSRNet, SASNet also uses the first ten layers of VGG-16 \cite{DBLP:journals/corr/SimonyanZ14a}, pre-trained on ImageNet \cite{5206848}, as feature extractor, and fuses features extracted by these layers across multiple scales to obtain an accurate density map.

\subsection{Gigapixel Deep Learning}

The term ``gigapixel'' suggests an image containing one billion pixels. However, images with resolutions ranging from 100 megapixels up to hundreds of gigapixels are considered to be ``gigapixel images'' in the literature \cite{bakhtiarnia2022efficient}.
Using gigapixel images and videos reveals much more detail about the scene and has the potential to significantly improve the accuracy of deep learning tasks. However, as previously mentioned, processing gigapixel images with deep learning is challenging due to GPU memory and computation limits. Even without considering GPU limits, existing deep learning architectures and methods are not capable of properly training the massive number of parameters that would result from using gigapixel images directly as input. Moreover, gigapixel datasets typically contain a very low number of images, since manually labelling such large images is a difficult task. For instance, the PANDA dataset contains only 45 examples compared to the 1,535 examples UCF-QNRF.

\begin{figure*}
\setlength{\fboxrule}{0pt}
\begin{center}
\begin{tabular}{ c c }
\includegraphics[width=0.48\textwidth]{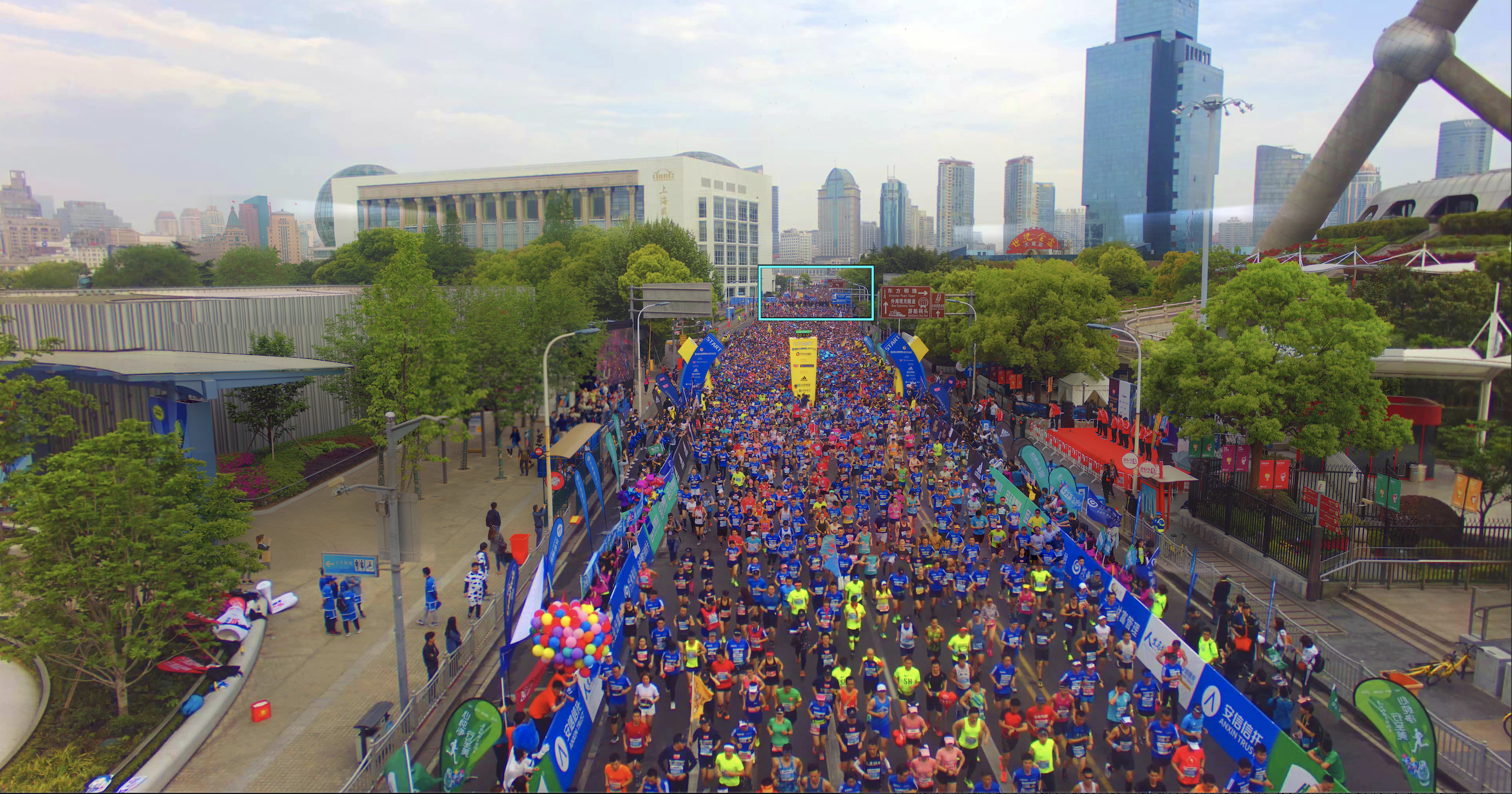} &
\includegraphics[width=0.48\textwidth]{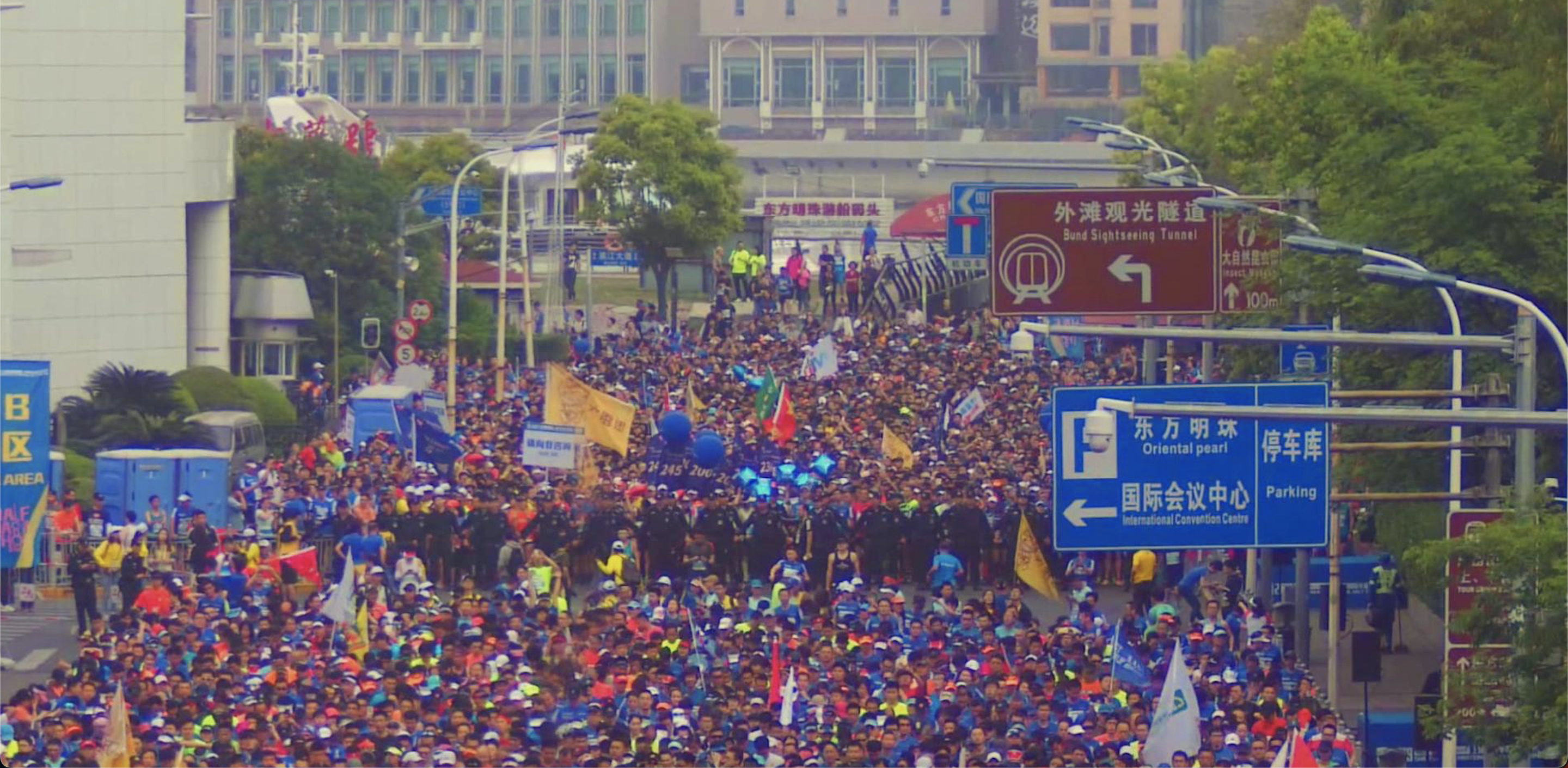}\\
\end{tabular}
\end{center}
\caption{(left) Example gigapixel image from the PANDA dataset, with a resolution of 26,908$\times$15,024 downsampled to 2,688$\times$1,412; and (right) zoomed into the region specified by the rectangle in the original image, with a resolution of 2,880$\times$1,410 pixels.}
\label{fig:example_gigapixel}
\end{figure*}

The most common approach for dealing with very high resolutions in deep learning is to downsample the images to a manageable resolution. However,
this obscures details and negates the benefits of capturing gigapixel images. For instance, as shown in Figure \ref{fig:example_gigapixel}, there are locations in the downsampled gigapixel image where several people are represented by a single pixel, making it impossible for a deep learning model to accurately predict crowd density.

Processing gigapixel whole-slide images (WSIs) is common in histopathology for cancer detection, detecting metastatis (the spread of cancer), neuropathology and detecting tissue components \cite{bakhtiarnia2022efficient}. For instance, HIPT \cite{Chen_2022_CVPR} processes gigapixel WSIs using a hierarchy of Vision Transformers, and \cite{8809829} uses neural image compression on WSIs so they can be processed with a CNN on a single GPU. However, a key difference between histopathology and crowd counting is the lack of perspective in the former. This means that in WSIs, cells and tissues always have roughly the same size, whereas in gigapixel crowd counting, the bounding box for a person near the camera can be up to 1 million times larger than that of a person far away.

Several methods exist for gigapixel object detection. For instance, GigaDet \cite{CHEN202214} is a near real-time object detection method for gigapixel videos. GigaDet counts the number of objects on regions of downsampled version of image across multiple scales, then processes the top candidate regions to detect objects. However, as explained in section \ref{sec:method}, gigapixel object detection methods cannot be directly used for crowd counting.

\section{GigaZoom}
\label{sec:method}

GigaZoom is inspired by how people act when they are asked to count the number of people in gigapixel images, where they zoom into the dense regions of the crowd until they can distinguish individuals. Similarly, GigaZoom iteratively zooms into multiple dense regions to refine the coarse density map. Section \ref{sec:zoom} provides the details of the zooming and refinement process, and section \ref{sec:cluster} describes how the multiple regions are detected.

\subsection{Iterative Zooming and Replacing}
\label{sec:zoom}

\begin{figure*}
\setlength{\fboxrule}{0pt}
\begin{center}
\includegraphics[width=\textwidth]{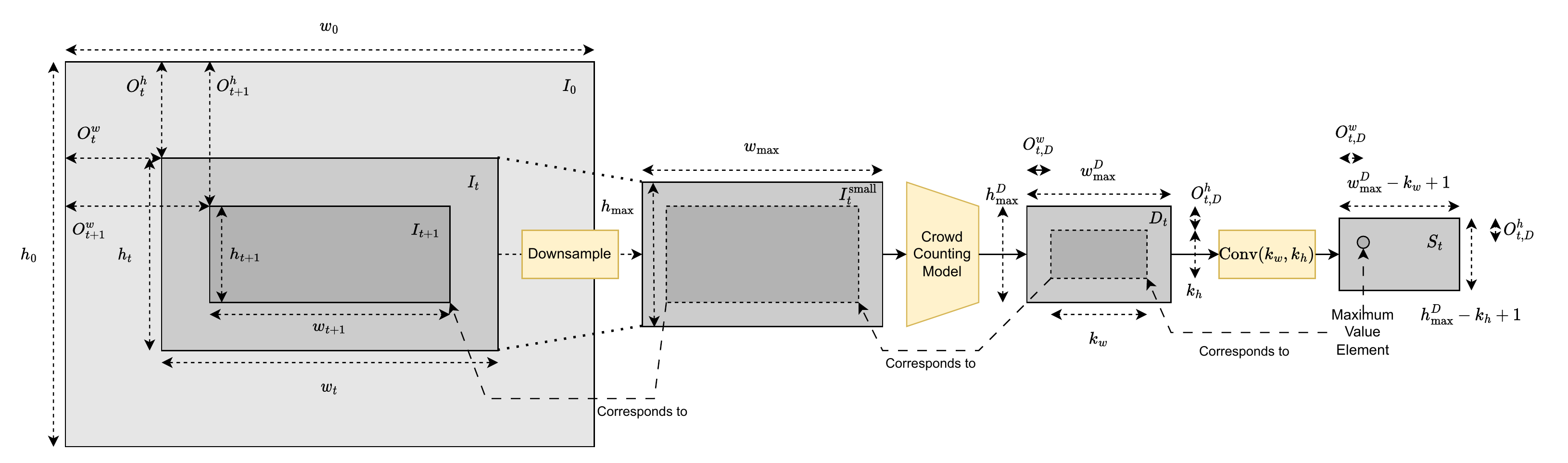}
\end{center}
\caption{Overview of the forward pass in iterative zooming and replacing.}
\label{fig:forward_pass}
\end{figure*}

Iterative zooming and replacing consists of two steps: a forward pass that iteratively zooms into the densest area of the image, and a backward pass that combines the density maps obtained during the forward pass to construct the final density map. Figure \ref{fig:forward_pass} shows an overview of the forward pass. Given a gigapixel image $ I_0 $ of resolution $ w_0 \times h_0 $, we perform $ L $ zoom-in operations until we reach a resolution within GPU memory limits. Note that $ L $ is a hyper-parameter of the method. The location of the zoomed-in image $ I_{t+1} $ depends on the density map obtained by previous image $ I_t $. Since resolution of $ I_t $ is beyond the GPU memory limit for $ t < L $, we are not able to use $ I_t $ directly as input to the crowd counting model. Therefore, we first need to downsample $ I_t $ to $ w_\text{max} \times h_\text{max} $, defined as the maximum image resolution that can fit into the available GPU memory. 

The width and height of $ I_t $ are determined based on the zoom formula. \textit{Linear zoom} is defined as
\begin{equation}
h_{t} = h_0 - \left(\frac{h_0 - h_\text{max}}{L}\right) t, \hspace{1cm} w_{t} = w_0 - \left(\frac{w_0 - w_\text{max}}{L}\right) t;
\label{linear_zoom}
\end{equation}
whereas \textit{exponential zoom} is defined as
\begin{equation}
h_{t} = h_0 \left(\frac{h_\text{max}}{h_0}\right)^\frac{t}{L}, \hspace{1cm} w_{t} = w_0 \left(\frac{w_\text{max}}{w_0}\right)^\frac{t}{L}.
\label{exponential_zoom}
\end{equation}

Suppose that we have performed $ t $ zoom-ins so far and obtained image $ I_t $ within $ I_0 $, where $ (O^w_t, O^h_t) $ is the top left corner of $ I_t $ inside $ I_0 $. Since the width and height of $ I_{t+1} $ are already known based on the zoom formula, our goal is to determine $ (O^w_{t+1}, O^h_{t+1}) $, which is the top left corner of $ I_{t+1} $ inside $ I_0 $. We start by uniformly downsampling $ I_t $ to $ I^\text{small}_t $ with a resolution of $ w_\text{max} \times h_\text{max} $. We then pass $ I^\text{small}_t $ to a crowd counting model to obtain density map $ D_t $. Note that the density map size $ w^D_\text{max} \times h^D_\text{max} $ might be smaller in size than $ I^\text{small}_t $ due to pooling operations in the crowd counting model.

The density of all sub-images within $ I_t $, which are candidates for $ I_{t+1} $, can be calculated using a simple convolution operation on the obtained density map, with an all-ones kernel of size $ k_w \times k_h $, where
\begin{equation}
k_w = \left(\frac{w_{t+1}}{w_t}\right) w^D_\text{max}, \hspace{1cm} k_h = \left(\frac{h_{t+1}}{h_t}\right) h^D_\text{max}.
\label{kernel_size}
\end{equation}
In the resulting matrix $ S_t $, the point $ (O^w_{t,D}, O^h_{t,D}) $ with the maximum value corresponds to the sub-image with the highest density. The top left corner of $ I_{t+1} $ can then be determined based on
\begin{equation}
\frac{O^w_{t+1} - O^w_t}{w_t} = \frac{O^w_{t,D}}{w^D_\text{max}}, \hspace{1cm} \frac{O^h_{t+1} - O^h_t}{h_t} = \frac{O^h_{t,D}}{h^D_\text{max}}.
\label{topleft}
\end{equation}

\begin{figure*}
\setlength{\fboxrule}{0pt}
\begin{center}
\includegraphics[width=0.95\textwidth]{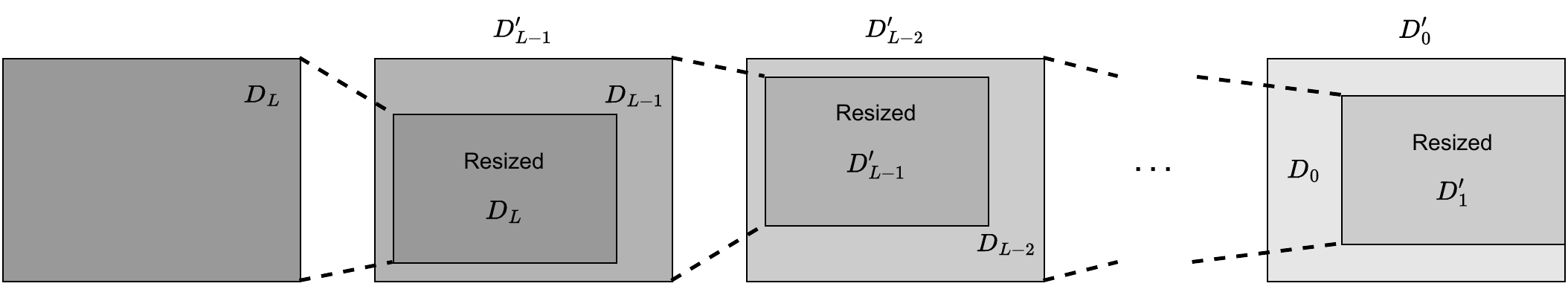}
\end{center}
\caption{Overview of the backward pass in iterative zooming and replacing.}
\label{fig:backward_pass}
\end{figure*}

Figure \ref{fig:backward_pass} shows an overview of the backward pass. During the forward pass, the density maps $ D_t, \:t=0, \dots, L $ along with the region of $ D_t $ that corresponds to $ D_{t+1} $ are saved to be used in the backward pass. The backward pass starts by resizing the finest density map $ D_L $ and replacing the region of $ D_{L-1} $ that corresponds to $ D_L $ to obtain an improved density map $ D'_{L-1} $. Subsequently, $ D'_{L-1} $ is resized and placed in the correponding region in $ D_{L-2} $, and this process is repeated until the final improved density map is obtained. We tested more complex merging operations than simply replacing regions of density maps, for instance, we trained a CNN to combine $ D_t $ and resized $ D'_{t+1} $ to obtain an estimation closer to the corresponding part of the ground truth density map. However, replacement always resulted in the highest accuracy. Another simple merging operation is averaging, which is used in Gigapixel CSRNet. However, taking the average of density maps across multiple scales is not sensible, since the more zoomed-in density maps are almost always more accurate.

Crowd counting models are designed and trained for a specific range of crowd density, therefore, if the density goes above or falls below that range, their error increases. Another advantage of GigaZoom over Gigapixel CSRNet is that by zooming into dense areas, it ensures that low density areas are not processed separately. In contrast, Gigapixel CSRNet always detects a small crowd of people even if the image patch is completely empty. This is exacerbated by the fact that in gigapixel images, many locations of the image are empty, resulting in a massive error. Note that empty regions are not an issue for gigapixel object detection methods, since they would simply be ignored. However, since, the density maps are added together in crowd counting, the errors accumulate.

\subsection{Multiple Zoom Regions}
\label{sec:cluster}

\begin{figure*}
\setlength{\fboxrule}{0pt}
\begin{center}
\begin{tabular}{ c c }
\includegraphics[width=0.57\textwidth]{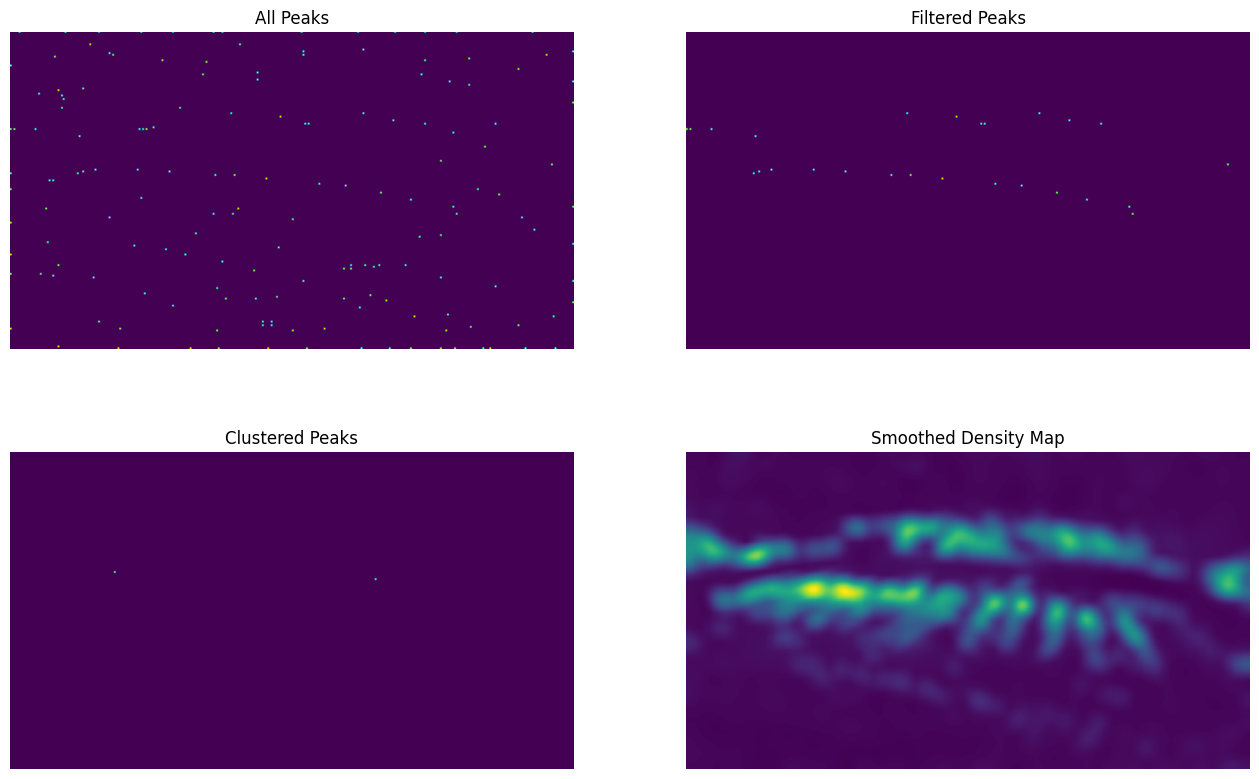} &
\includegraphics[width=0.38\textwidth]{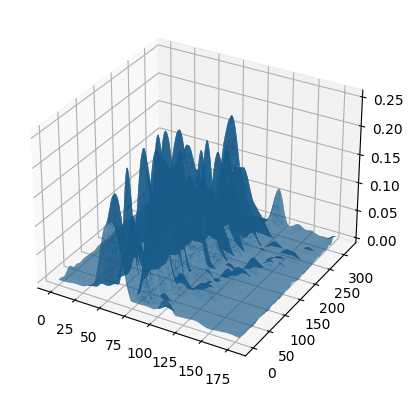}\\
(a)&(b)
\end{tabular}
\end{center}
\caption{(a) The smoothed density map as well as detected, filtered and clustered peaks; and (b) three-dimensional visualization of the smoothed density map.}
\label{fig:peaks}
\end{figure*}

Iterative zooming and replacing only zooms into a single region. However, multiple dense regions might be present in a given image. Therefore, we specify several regions to apply iterative zooming and replacing. We start by smoothing the coarsest density map $ D_0 $ using a Gaussian filter to remove small spikes in density. Peaks in the smoothed density map are then detected using a local maximum filter \cite{WULDER2000103}. The detected peaks are then filtered based on a threshold $ \lambda $, and the remaining peaks are clustered using the k-means algorithm to $ k $ clusters. Finally, we apply iterative zooming and replacing on sub-images centered at the cluster centers. The overall process is depicted in Figure \ref{fig:peaks}.

Note that using multiple zoom regions may lead to conflicts since some areas might be processed during several iterative zooming and replacing operations. To resolve these conflicts, we tested several aggregation strategies such as averaging or using the maximum value. However, we found that all strategies obtain similar results. Therefore, we opted for the simplest strategy, which is to use the latest result in case of a conflict.

GigaZoom performs $ k \times L $ CSRNet inferences per input gigapixel image. With the hyper-parameters specified in section \ref{sec:experiments}, this translates to 20 CSRNet inferences, which is more than 10$\times$ faster than Gigapixel CSRNet, which performs an average of 204 CSRNet inferences per input gigapixel image. Note that iterative zooming and replacing cannot be parallelized, however, multiple iterative zooming and replacing operations can be performed in parallel.

\section{Experiments}
\label{sec:experiments}

\subsection{Setup and Results}

Since PANDA \cite{Wang_2020_CVPR} does not specify training and test splits, we selected 30 images for training, 6 images for validation and 9 images for test. The selection procedure can be viewed in our code. To obtain a ground truth density map from the bounding box annotations available in the PANDA dataset, for each bounding box, we apply a 2D Gaussian filter with $ \sigma = 4 $ and filter size the same as the bounding box.

The hyper-parameters used in GigaDet are as follows. We use exponential zoom with $ L=10 $. The maximum resolution $ w_\text{max} \times h_\text{max} $ that fits our GPU memory is 2,560$\times$1,440. To determine multiple zoom regions, a Gaussian filter with $ \sigma = 4 $ and radius of 7 is used for smoothing, threshold $ \lambda = 0.1 $ is used for filtering and number of clusters $ k = 2 $ is used for clustering. We use two separate crowd counting models: a PromptMix model \cite{bakhtiarnia2023promptmix} to obtain $ D_0 $, and for all other density maps $ D_1, \dots, D_L $ we use a CSRNet model \cite{cao2019gigapixel} trained on patches of different scales. The first model is trained with the procedure outlined in \cite{bakhtiarnia2023promptmix}, and the second model is trained by initializing with pre-trained weights from the PromptMix model, and fine-tuning for 100 epochs with a weight decay of $ 10^{-4} $, batch size of 12 and a learning rate of $ 10^{-4} $ which is multiplied by 0.99 each epoch. All experiments were conducted on 3$ \times $Nvidia A6000 GPUs, each with 48 GBs of video memory.

Crowd counting methods are typically evaluated by using the mean absolute error (MAE) or the mean squared error (MSE), defined as
\begin{equation}
\text{MAE} = \frac{\sum^N_{i=1}|\hat{y}_i - y_i|}{N}, \hspace{1cm} \text{MSE} = \frac{\sum^N_{i=1}(\hat{y}_i - y_i)^2}{N},
\label{mae_mse}
\end{equation}
where $ \hat{y}_i $ is the prediction for $i$-th image, $ y_i $ is the ground truth label, and $ N $ is the total number of examples in the dataset. In crowd counting, MAE is typically used as a measure of accuracy, whereas MSE is a measure of robustness \cite{10.1145/3460426.3463628}. Since our primary objective is accuracy, we use MAE to evaluate crowd counting methods in this work.

The original SASNet paper does not include experiments on the PANDA dataset \cite{Song_Wang_Wang_Tai_Wang_Li_Wu_Ma_2021}, therefore, we initalize the training with pre-trained weights for Shanghai Tech Part A and fine-tune on the PANDA dataset downsampled to 2,560$\times$1,440 pixels. Although Gigapixel CSRNet uses the PANDA dataset, the authors do not report accuracy metrics, therefore, we reproduce the method to measure its accuracy. Since PromptMix includes experiments on PANDA, we use the number from the original paper. Table \ref{tab:results} compares the accuracy of GigaZoom with previous methods on the PANDA dataset. Observe that GigaZoom significantly outperforms other methods.

\begin{table}
\begin{center}
\begin{tabular}{| l l l |}
\hline
Method & Year & MAE$\downarrow$\\
\hline
\hline
Gigapixel CSRNet \cite{cao2019gigapixel} & 2019 & 2680.20\\
SASNet \cite{Song_Wang_Wang_Tai_Wang_Li_Wu_Ma_2021} & 2021 & 263.88\\
PromptMix \cite{bakhtiarnia2023promptmix} & 2023 & 110.34\\
GigaZoom (ours) & 2023 & \textbf{63.51}\\
\hline
\end{tabular}
\end{center}
\caption{Comparison of crowd counting performance for various methods on the PANDA gigapixel dataset.}
\label{tab:results}
\end{table}

\subsection{Ablation Studies}

Table \ref{tab:ablation_zoom} compares the accuracy obtained by the two different zoom methods defined in Equations \ref{linear_zoom} and \ref{exponential_zoom}, which shows that exponential zoom leads to a higher accuracy. Table \ref{tab:ablation_levels} shows the effect of the number of zoom levels $L$ on the accuracy. Based on these results, using too few or too many zoom levels can lead to sub-optimal accuracy. Table \ref{tab:ablation_clustering} shows that using multiple zoom regions can slightly boost the accuracy. However, similar to the number of zoom levels, using too few or too many clusters can degrade the accuracy. Even though the accuracy improvement is slight in these experiments, using multiple zoom regions makes GigaZoom more robust and might lead to more significant improvements is other scenarios and scenes. We also investigated the effect of \textit{overzooming} in Table \ref{tab:ablation_overzooming}. Overzooming is defined as zooming beyond a 1-to-1 pixel ratio, where several pixels in the resulting image correspond to a single pixel in the original image, effectively upsampling a region of the original gigapixel image. However, these results show that the method does not benefit from overzooming.

\begin{table}
\begin{center}
\begin{tabular}{| l l |}
\hline
Zoom Method & MAE$\downarrow$\\
\hline
\hline
Linear & 81.02\\
Exponential & 63.51\\
\hline
\end{tabular}
\end{center}
\caption{Effect of linear and exponential zoom on the accuracy of GigaZoom.}
\label{tab:ablation_zoom}
\end{table}

\begin{table}
\begin{center}
\begin{tabular}{| c l |}
\hline
Zoom Levels ($ L $) & MAE$\downarrow$\\
\hline
\hline
5 & 80.04\\
10 & 64.49\\
20 & 105.45\\
\hline
\end{tabular}
\end{center}
\caption{Effect of the number of zoom levels on the accuracy of GigaZoom. Clustering was not used in these experiments, and only a single pass of iterative zooming and replacing was performed on the densest sub-image of the input.}
\label{tab:ablation_levels}
\end{table}

\begin{table}
\begin{center}
\begin{tabular}{| c c l |}
\hline
Multiple Zoom Regions & Clusters & MAE$\downarrow$\\
\hline
\hline
$\times$ & - & 64.49\\
\checkmark & 1 & 70.19\\
\checkmark & 2 & 63.51\\
\checkmark & 5 & 79.34\\
\hline
\end{tabular}
\end{center}
\caption{Effect of multiple zoom regions on the accuracy of GigaZoom.}
\label{tab:ablation_clustering}
\end{table}

\begin{table}
\begin{center}
\begin{tabular}{| c c l |}
\hline
Zoom Levels ($L$) & Overzoom Levels & MAE$\downarrow$\\
\hline
\hline
10 & 0 & 64.49\\
10 & 1 & 67.97\\
10 & 2 & 93.26\\
\hline
\end{tabular}
\end{center}
\caption{Effect of overzooming on the accuracy of GigaZoom. Clustering was not used in these experiments, and only a single pass of iterative zooming and replacing was performed on the densest sub-image of the input.}
\label{tab:ablation_overzooming}
\end{table}

\section{Conclusion}
\label{sec:conclusion}

We showed that our proposed method signifcantly outperforms existing methods for crowd counting on gigapixel images. Through ablation studies, we showed that exponential zoom performs better than linear, a moderate number of zoom levels achieves best accuracy, and using multiple zoom regions provides robustness for inputs with multiple dense crowds. Although GigaZoom is much more efficient than Gigapixel CSRNet, it still performs multiple CSRNet inferences, which can result in a long inference time overall.

Currently, PANDA is the only publicly available dataset for this task, which contains only 45 images taken from three scenes. In order to further compare and validate methods, it is crucial that more gigapixel crowd counting datasets are created and published, that have a higher number of examples taken from more diverse scenes.

\bibliographystyle{IEEEtran}
\bibliography{references.bib}

\end{document}